\documentclass[conference]{IEEEtran}
\usepackage[hyphens]{url}
\usepackage{hyperref}
\hypersetup{breaklinks=true,colorlinks,allcolors=blue}
\usepackage{doi}

\IEEEoverridecommandlockouts
\usepackage{cite}
\usepackage{caption}
\usepackage{algorithmic}
\usepackage{graphicx}
\usepackage{textcomp}
\usepackage{xcolor}
\usepackage{amsmath}
\usepackage{balance}
\usepackage[left=1.62cm,right=1.62cm,top=1.9cm]{geometry}
\usepackage{graphicx}
\usepackage{subcaption}
\usepackage{makecell}

\def\BibTeX{{\rm B\kern-.05em{\sc i\kern-.025em b}\kern-.08em
    T\kern-.1667em\lower.7ex\hbox{E}\kern-.125emX}}
\begin{document}

\newcommand{\james}[1]{{\color{magenta}James: #1}}
\newcommand{\ahmed}[1]{\textcolor{magenta}{Dr. Ahmed: {#1}}}


\title{Synthetic Thermal Image Generation for Real-Time Animal Detection Under Low-Visibility Conditions}

\author{\IEEEauthorblockN{James Momoh and Khandaker Mamun Ahmed$^*$}
 \IEEEauthorblockA{The Beacom College of Computer and Cyber Sciences, Dakota State University \\
Email: James.Momoh@trojans.dsu.edu, khandakermamun.ahmed@dsu.edu}
}

\maketitle
\let\thefootnote\relax\footnote{$^*$Corresponding Author.}

\begin{abstract}
Wildlife–vehicle collisions remain a significant road safety concern, particularly during nighttime and low-visibility conditions when RGB-based perception systems are often unreliable. Thermal imaging offers a promising alternative for detecting animals under poor illumination. However, the limited availability of annotated infrared animal datasets restricts the development of robust deep learning-based detection models. This paper investigates synthetic thermal image generation as a scalable approach for real-time animal detection under low-visibility conditions. A subset of 514 annotated visible-spectrum animal images from the NTLNP dataset is translated into synthetic thermal representations using CycleGAN-Turbo, while a limited real thermal dataset of 60 images is expanded through thermal-focused augmentation. Multiple object detection architectures, including YOLOv8, YOLOv9, YOLOv10, and RT-DETR, are trained independently on synthetic and real thermal datasets and evaluated using precision, recall, mAP@0.5, mAP@0.5:0.95, model size, and inference latency. Experimental results show that synthetic thermal images provide competitive detection performance, with RT-DETR achieving the highest synthetic-data mAP@0.5 of 0.9613. Models trained on augmented real thermal data achieve the strongest overall performance, with YOLOv10s obtaining 0.9879 mAP@0.5 and 0.9571 mAP@0.5:0.95. Computational analysis further indicates that lightweight YOLO variants provide favorable inference latency, supporting their potential for real-time deployment. These findings demonstrate that synthetic thermal imagery can reduce dependence on scarce infrared datasets and support the development of efficient animal detection systems for future vehicle-mounted wildlife collision mitigation applications.
\end{abstract}


\section{Introduction}
Wildlife–vehicle collisions (WVCs) remain a significant and persistent safety challenge across the United States, with substantial human, ecological, and economic consequences. Over one million wildlife–vehicle collisions occur annually in the U.S., resulting in hundreds of human fatalities, tens of thousands of injuries, and billions of dollars in damages each year \cite{fhwa_wvc_2008}. More recent estimates further highlight the scale of the problem, suggesting that the total annual cost of WVCs exceeds \$8 billion when accounting for vehicle damage, medical expenses, and insurance claims \cite{statefarm_wvc_2023}. Similarly, these collisions are both widespread and costly, with increasing trends observed due to expanding road networks and wildlife habitat fragmentation \cite{pew_wvc_2021}.

A major contributing factor to wildlife–vehicle collisions is reduced visibility, particularly during nighttime or low-light conditions when many animals are most active. Studies indicate that the majority of WVCs occur during dusk, dawn, or nighttime periods, when driver perception is limited and reaction time is reduced \cite{fhwa_wvc_2008, statefarm_wvc_2023}. Traditional vision-based detection systems, which rely on visible spectrum (RGB) cameras, are inherently constrained under these conditions due to poor illumination, glare, and environmental occlusions. As a result, these systems may not detect animals early enough to prevent collisions, limiting their effectiveness in real-world deployments. 

Thermal imaging presents a promising alternative for addressing visibility challenges in such environments. Unlike RGB imaging, thermal cameras detect heat signatures emitted by objects, enabling reliable identification of animals regardless of lighting conditions. This capability makes thermal imaging suitable for safety-critical applications such as driver assistance systems and autonomous vehicle perception. Prior research in autonomous driving has demonstrated that thermal cameras provide robust perception capabilities in darkness and challenging environmental conditions where RGB-based systems fail \cite{mazhr2024thermal}. Similarly, studies on RGB-thermal fusion have shown that thermal data can complement visible imagery and significantly improve detection robustness in low-visibility scenarios \cite{sun2021fuseseg}, \cite{hazra2025crossmodal}.

Despite these advantages, the development of thermal-based detection systems remains limited by the scarcity of large-scale annotated infrared datasets. Unlike RGB data, infrared imagery is costly to collect and annotate due to specialized sensor requirements, making data scarcity a key bottleneck in thermal object detection and RGB-to-thermal learning \cite{chen2025pseudoir,xiao2026thermalgen}. Therefore, to reduce reliance on real infrared data, recent studies have explored image-to-image translation methods that generate thermal-like images from readily available RGB inputs. These methods offer a scalable way to expand thermal training data. However, synthetic image generation alone does not guarantee improved detection performance. Generated thermal images must preserve the structural and semantic cues required for object localization, and their usefulness must be validated through downstream detection tasks. Moreover, prior work has often treated thermal image generation, detection accuracy, and deployment efficiency as separate problems, leaving limited understanding of how synthetic thermal data performs for possible real-time animal detection under low-visibility conditions.

This study addresses this gap by evaluating synthetic thermal image generation as a practical training strategy for animal detection. 
Unlike existing studies that primarily focus on either RGB-to-thermal image translation or thermal object detection independently, this work provides a unified evaluation framework that investigates the effectiveness of synthetic thermal imagery for wildlife detection under low-visibility conditions.
Specifically, RGB animal images are translated into thermal-like representations and used to train state-of-the-art object detection models. Their performance is compared with models trained on augmented real thermal imagery to assess whether synthetic thermal data can serve as an alternative to scarce infrared data. In addition, lightweight YOLO and RT-DETR architectures are evaluated using both detection accuracy and computational efficiency metrics, including inference latency, and parameter size to examine their suitability for future vehicle-mounted deployment.

The contributions of this work are summarized as follows:

\begin{itemize}
    \item A synthetic thermal animal dataset is generated by translating 514 visible-spectrum animal images from the NTLNP dataset into infrared-like representations.
    \item The effectiveness of synthetic thermal images is evaluated as an alternative to real infrared data for training animal detection models.
    \item Lightweight YOLO and RT-DETR detection architectures are evaluated in terms of detection accuracy and computational efficiency to assess their suitability for real-time, vehicle-mounted animal detection aimed at mitigating wildlife-vehicle collisions.
\end{itemize}
\section{Related Work} \label{related_works}

\subsection{Thermal Imaging for Low-Visibility Perception}

Thermal imaging has been widely studied as a robust sensing modality for perception in low-light and adverse environmental conditions. Unlike RGB cameras, thermal sensors capture emitted heat signatures, enabling consistent performance regardless of illumination. In autonomous driving research, thermal imaging has been shown to enhance object detection and scene understanding in nighttime and challenging environments \cite{mazhr2024thermal}.

To further improve perception, several studies have explored the fusion of RGB and thermal data. For instance, FuseSeg introduced an end-to-end deep learning framework that combines RGB and thermal inputs to improve semantic segmentation in urban scenes, particularly under poor lighting conditions \cite{sun2021fuseseg}. Similarly, cross-modal attention-based fusion approaches have demonstrated improved robustness and accuracy by leveraging complementary information from both modalities \cite{hazra2025crossmodal}. While these methods improve detection performance, they typically rely on the availability of both RGB and thermal data, which may not always be feasible in practice.

\subsection{Infrared Object Detection}

Infrared object detection has attracted significant attention due to its applications in autonomous driving, surveillance, and wildlife monitoring. However, detecting objects in infrared images presents unique challenges, including low resolution, limited texture information, and small object sizes. Prior surveys highlight that while infrared detection is effective in harsh environments, it requires specialized techniques to overcome these limitations \cite{huo2022infrared}. Therefore, recent approaches have focused on improving detection performance through architectural innovations. Attention mechanisms, multi-scale feature fusion, and transformer-based models have been proposed to enhance feature representation and improve detection accuracy. For example, layered transformer networks have been used to enhance contrast and improve multi-object detection in thermal imagery \cite{li2024infrared_transformer}. Similarly, multi-scale upsampling and spatial attention techniques have been integrated into YOLO-based models to improve detection of small infrared objects \cite{cai2024infrared}.

There has also been a growing emphasis on lightweight models for real-time applications. Optimized versions of RT-DETR and YOLO have been developed to reduce computational complexity while maintaining high accuracy, making them suitable for deployment in resource-constrained environments such as UAVs and embedded systems \cite{du2024rtdetr}, \cite{liu2025disodetr}. 

\subsection{Visible-to-Infrared Image Translation}
To address the scarcity of infrared datasets, researchers have explored image translation techniques that convert visible images into infrared-like representations. CycleGAN remains a foundational approach in this domain, enabling unpaired image-to-image translation through cycle-consistency constraints \cite{zhu2017cyclegan}. Subsequent works have extended this framework to support few-shot learning and improved diversity in generated images \cite{liu2019fewshot}, \cite{lee2018drit}.
In the context of object detection, recent studies have investigated the use of synthetic infrared images for training detection models. The authors in \cite{chen2025pseudoir} demonstrated that pseudo-infrared images generated using CycleGAN can improve the performance of YOLOv8-based detectors when evaluated on real infrared datasets. Similarly, authors in \cite{park2025vida} proposed an unsupervised domain adaptation framework that generates infrared images from RGB inputs, enabling detection models to be trained without direct infrared data.
These approaches highlight the potential of synthetic data as a scalable alternative to real infrared datasets. However, challenges remain in ensuring that generated images accurately preserve the semantic and structural characteristics necessary for reliable detection.

\subsection{ Detection and Edge Deployment}
Wildlife detection has been studied in both ecological monitoring and road safety contexts. Early work demonstrated the use of thermal imaging for detecting animals using UAV systems, highlighting its effectiveness in identifying heat signatures in natural environments \cite{ward2016uav_wildlife}. More recent studies have applied deep learning models such as YOLO and transformer-based architectures for detecting animals in aerial imagery, achieving high accuracy in challenging conditions \cite{roca2025deer}, \cite{siddique2023deer}.
In parallel, there has been increasing interest in deploying detection models on edge devices for real-time applications. Lightweight architectures, including optimized YOLO variants and transformer-based models, have been implemented on embedded platforms such as Raspberry Pi and Jetson Nano, demonstrating the feasibility of real-time wildlife detection in resource-constrained environments \cite{katangure2025wildlife}.
Despite these advancements, existing studies often treat thermal perception, synthetic data generation, and wildlife detection as separate problems. There remains a gap in integrating these components into a unified framework for real-time animal detection under low-visibility conditions. This study addresses this gap by combining synthetic thermal generation, efficient detection architectures, and deployment-aware design into a unified evaluation framework.

\section{Proposed Method} \label{System Design and Implementation}

\begin{figure}[!ht]
    \centering
    \includegraphics[width=\linewidth]{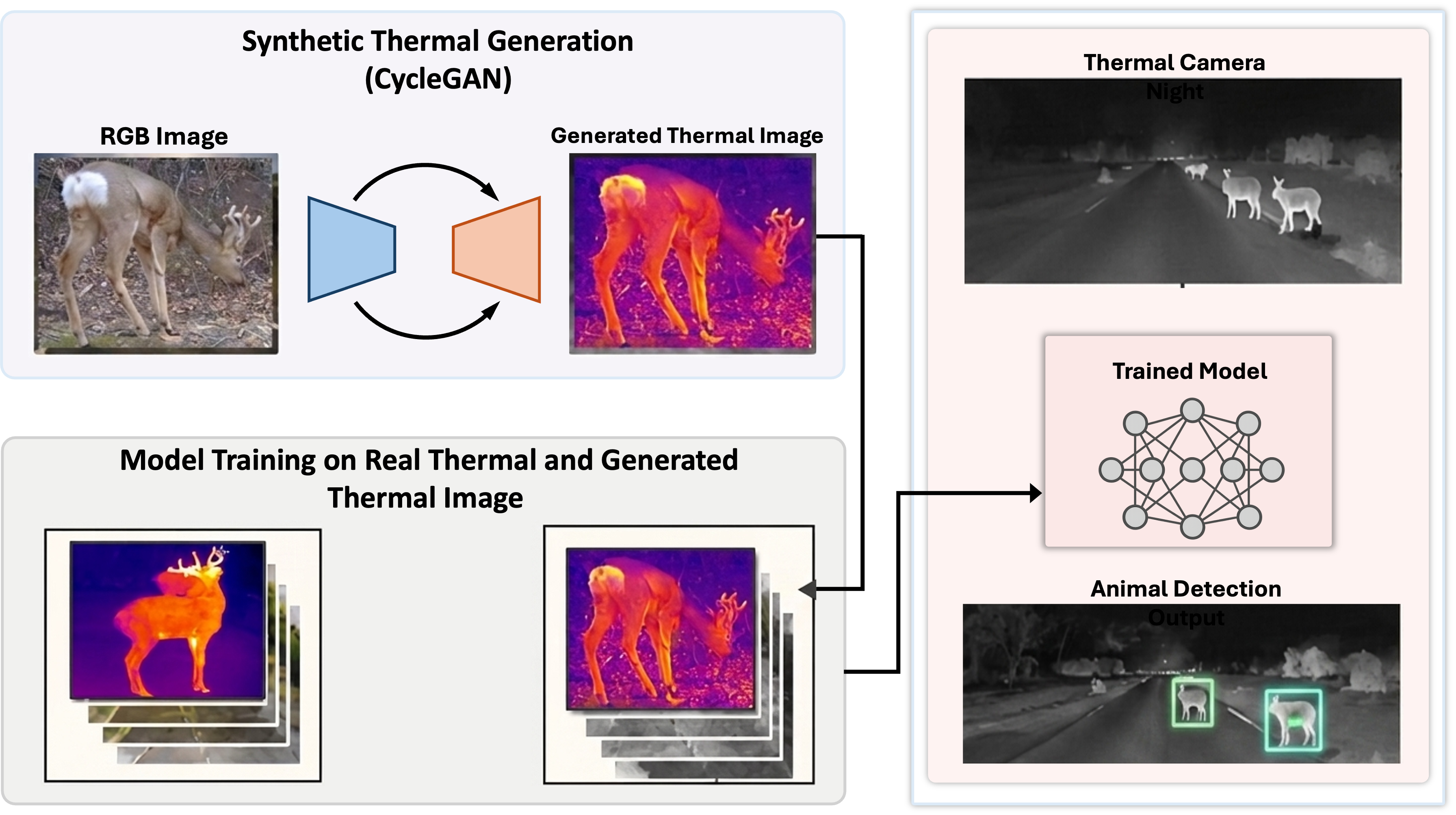}
    \caption{Proposed thermal animal detection framework. RGB wildlife images are translated into synthetic thermal representations using CycleGAN-Turbo. Object detection models are trained independently on synthetic and augmented real thermal datasets and evaluated for detection accuracy and computational efficiency.}
    \label{fig:framework}
\end{figure}

\subsection{Problem Formulation} \label{agent_design}


Let the thermal domain be denoted by $T$, consisting of two distinct data sources:
\begin{itemize}
    \item Synthetic thermal images $\tilde{x}_t$, generated from RGB images
    \item Real thermal images $x_t$, collected from limited infrared datasets
\end{itemize}

Each image is associated with object detection annotations:
\begin{equation}
y = \{c, b_x, b_y, w, h\}
\end{equation}

where $c$ is the class label and $(b_x, b_y, w, h)$ represent bounding box parameters.

This study adopts a training strategy in which models are trained independently on:

\[
D_{\text{syn}} = \{(\tilde{x}_t, y)\}, \quad
D_{\text{real}} = \{(x_t^{\text{aug}}, y)\}
\]

This design enables a controlled evaluation of the effectiveness of synthetic thermal data as a substitute for real infrared data. The goal is to improve or at least maintain animal detection performance under low-visibility conditions while minimizing reliance on large-scale real thermal datasets. The overall workflow of the proposed thermal animal detection framework is illustrated in Fig. \ref{fig:framework}.

\subsection{RGB-to-Thermal Image Translation}

To address the scarcity of real thermal data, synthetic thermal images are generated from RGB images using a CycleGAN-based image translation framework, following unpaired image-to-image translation principles.
Given an RGB image $x_s \in S$, a generator $G: S \rightarrow T$ produces a synthetic thermal image defined as:
\[
\tilde{x}_t = G(x_s;c)
\]

where $\tilde{x}_t$ represents the translated thermal image and $c$ denotes the target thermal-domain prompt used for conditional generation.

Unlike conventional CycleGAN implementations that rely on ResNet-based generators, this study adopts CycleGAN-Turbo which is built upon the Stable Diffusion Turbo (SD-Turbo) backbone. The framework integrates diffusion-based image translation with adversarial learning to improve thermal realism while preserving structural consistency.
The generator consists of domain-conditioned variational autoencoder (VAE) branches for RGB-to-thermal and thermal-to-RGB translation together with a shared text-conditioned UNet2DConditionModel. Low-Rank Adaptation (LoRA) layers are incorporated during fine-tuning to improve domain adaptation efficiency. Text conditioning is provided through a frozen CLIP text encoder using predefined prompts corresponding to the source (``rgb images") and target domains (``thermal images").

Adversarial supervision is provided through two vision-aided discriminators, one for each image domain, using a multilevel sigmoid GAN objective to distinguish translated thermal images from real thermal samples. This adversarial setup encourages the generated outputs to align closely with real infrared image distributions.
The translation model is trained using unpaired RGB and thermal datasets organized into separate domains. During training, images are resized to $146 \times 146$, randomly cropped to $128 \times 128$, horizontally flipped, normalized to the range $[-1,1]$, and converted into tensor representations. During inference, images are processed at their native resolution and resized back to the original spatial dimensions after translation.

Optimization is performed using AdamW optimizer with $\beta_1 = 0.9$ and  $\beta_2 = 0.999$, a weight decay of $10^{-2}$, and a learning rate of $10^{-5}$. Training is conducted for 100 epochs using a batch size of 1, corresponding to approximately 25,000 optimization iterations over the course of training. Gradient checkpointing is enabled to reduce memory consumption, and all experiments are executed using \texttt{float32} precision with a fixed random seed of 42.
The overall objective function combines adversarial, cycle-consistency, identity, and perceptual losses to balance realism and structural preservation. Cycle consistency and identity preservation are enforced using both $L_1$ and LPIPS perceptual losses, while adversarial learning is controlled using a GAN loss term. Gradient clipping with a maximum norm of 10 ($\|\nabla \theta\|_2 \leq 10$) is applied to stabilize training.

This translation process ensures that the generated synthetic thermal images preserve important object-level features such as animal shape, contours, and localization-relevant structures while adapting the visual appearance to thermal characteristics including monochromatic intensity distributions, reduced texture detail, and heat-emission-like representations.
The resulting synthetic thermal dataset provides a scalable alternative to real infrared imagery and forms the basis for evaluating the effectiveness of synthetic data in training object detection models under low-visibility conditions.



\subsection{Real Thermal Augmentation}

Due to the limited size of the real animal thermal dataset, augmentation is applied to enhance the data diversity and improve model generalization. The original real thermal dataset consists of approximately 60 thermal images collected from publicly available sources. To increase the effective training size, a thermal-focused augmentation strategy is employed with an augmentation factor of 4. Multiple augmentation operations are applied to generate augmented thermal samples $x_t^{\text{aug}}$. The selected transformations are designed to preserve and simulate thermal imaging characteristics.

The augmentation techniques include affine and perspective transformations to simulate viewpoint variation and camera distortion, horizontal flipping to improve orientation invariance, and brightness and contrast jittering to model varying thermal intensity distributions. Gamma correction and CLAHE (Contrast Limited Adaptive Histogram Equalization) are applied to enhance low-contrast thermal regions. Noise and blur simulation, including Gaussian noise, Gaussian blur, and motion blur, are also incorporated to imitate sensor noise and motion artifacts commonly observed in thermal cameras. Additionally, coarse dropout augmentation is employed to simulate partial occlusion and missing thermal regions during detection.

The augmentation process enables the detector to learn more robust thermal-domain representations under varying environmental and sensor conditions.
The resulting augmented dataset provides a benchmark for comparison against models trained on synthetic thermal data.

\section{Experimental Setup} \label{Experimental Setup}

\subsection{Dataset}
To evaluate the effectiveness of the proposed approach, two distinct datasets are constructed according to the training strategy.

\begin{figure}[!ht]
    \centering

    \begin{subfigure}[t]{0.48\linewidth}
        \centering

        \begin{subfigure}[b]{0.48\linewidth}
            \centering
            \includegraphics[width=\linewidth,keepaspectratio]{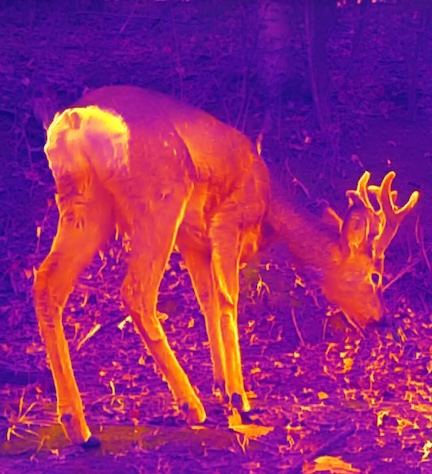}
        \end{subfigure}
        \hfill
        \begin{subfigure}[b]{0.48\linewidth}
            \centering
            \includegraphics[width=\linewidth,keepaspectratio]{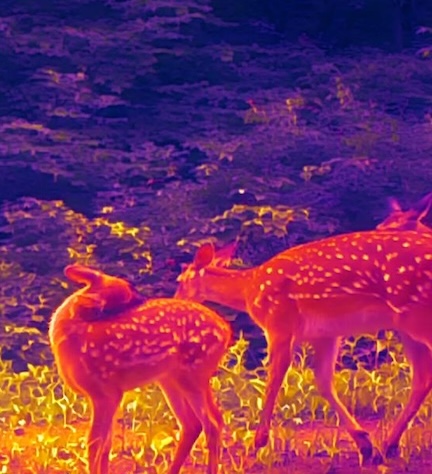}
        \end{subfigure}

        \vspace{0.1cm}

        \begin{subfigure}[b]{0.48\linewidth}
            \centering
            \includegraphics[width=\linewidth,keepaspectratio]{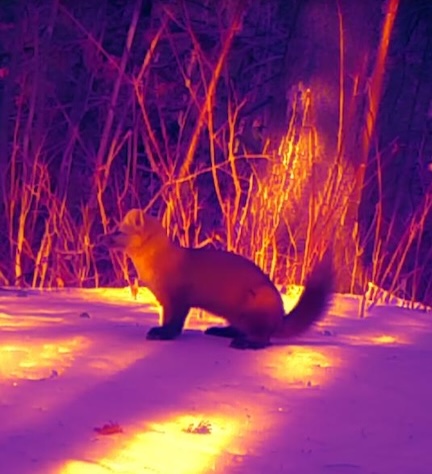}
        \end{subfigure}
        \hfill
        \begin{subfigure}[b]{0.48\linewidth}
            \centering
            \includegraphics[width=\linewidth,keepaspectratio]{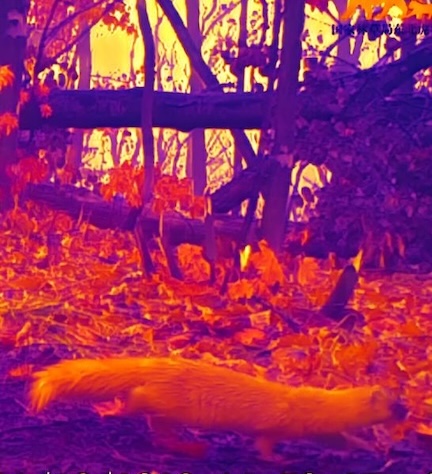}
        \end{subfigure}

        \caption{Synthetic thermal images}
        \label{fig:a}
    \end{subfigure}
    \hfill
    \begin{subfigure}[t]{0.48\linewidth}
        \centering

        \begin{subfigure}[b]{0.48\linewidth}
            \centering
            \includegraphics[width=\linewidth,keepaspectratio]{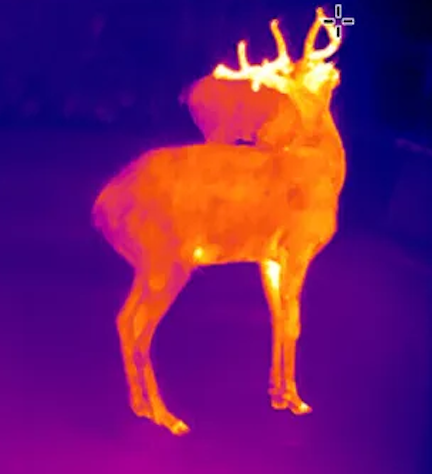}
        \end{subfigure}
        \hfill
        \begin{subfigure}[b]{0.48\linewidth}
            \centering
            \includegraphics[width=\linewidth,keepaspectratio]{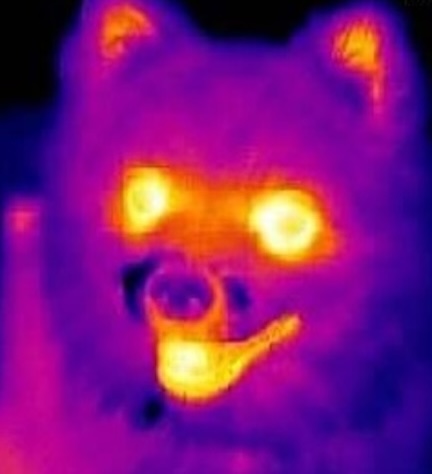}
        \end{subfigure}

        \vspace{0.1cm}

        \begin{subfigure}[b]{0.48\linewidth}
            \centering
            \includegraphics[width=\linewidth,keepaspectratio]{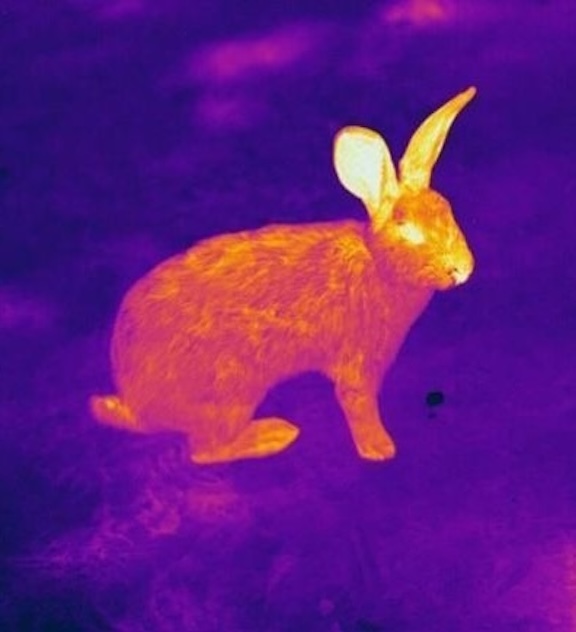}
        \end{subfigure}
        \hfill
        \begin{subfigure}[b]{0.48\linewidth}
            \centering
            \includegraphics[width=\linewidth,keepaspectratio]{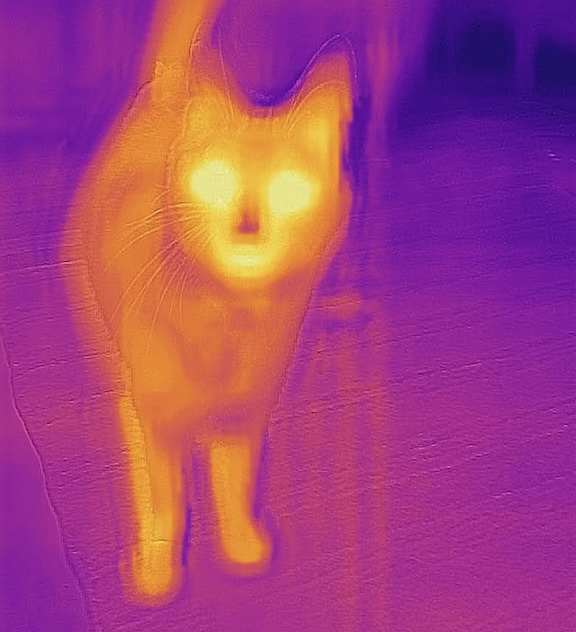}
        \end{subfigure}

        \caption{Real thermal images}
        \label{fig:b}
    \end{subfigure}

    \caption{Examples of synthetic and real thermal animal images. 
    (a) Synthetic thermal images generated from RGB images; 
    (b) Real thermal images collected from publicly available sources.}
    
    \label{fig:thermalcomparison}
\end{figure}

\textbf{Synthetic Thermal Dataset ($D_{syn}$).} Synthetic thermal images are generated from RGB images obtained from the NTLNP dataset using the RGB-to-thermal translation model. Examples of the generated synthetic thermal images are shown in Fig. \ref{fig:a}. Although the complete NTLNP dataset contains 15,314 visible-spectrum wildlife images, only 514 annotated images were selected and translated into synthetic thermal representations. The selected classes represent animals commonly found in the Midwestern United States and were chosen to support wildlife-vehicle collision detection under low-visibility conditions. The generated images preserve object structure while adapting to thermal-like intensity distributions, and the corresponding annotations are retained from the original RGB dataset. The synthetic dataset contains three animal classes: deer, sable, and weasel. These classes were selected to provide variation in animal size and appearance, allowing the detection models to be evaluated across different object scales. The synthetic thermal dataset was divided using an 80:20 train-validation split, resulting in 408 training images and 106 validation images.

\textbf{Real Thermal Dataset ($D_{real}$).} A limited set of real thermal images (60 samples) was collected from publicly available infrared sources. Representative examples of the real thermal images are shown in Fig. \ref{fig:b}. The dataset contains six animal classes: deer, cat, cow, boar, dog, and hare, representing a mixture of small- and large-sized animals commonly encountered in wildlife and rural environments. Due to the limited dataset size, the augmentation method is applied to generate an expanded training set.

\[
D_{\text{real}} = \{(x_t^{\text{aug}}, y)\}
\]

Rather than combining both datasets during training, models are trained independently on each dataset to enable a systematic comparison between synthetic and real thermal representations.

All experiments are conducted on a GPU-enabled HPC cluster using the Ultralytics framework (v8.4.7) with PyTorch-based implementations. Training jobs are executed through a SLURM-managed cluster to support large-scale benchmarking across multiple model variants. The training and inference experiments are performed using PyTorch 2.5.1 with CUDA 12.1 support on NVIDIA A100 GPUs equipped with 80 GB of memory. To ensure fairness and consistency, identical training configurations are maintained across experiments. Models are trained for up to 100 epochs, with the best-performing checkpoint selected based on validation mAP. The study evaluates lightweight and real-time detection architectures, including YOLOv8n, YOLOv8s, YOLOv9t, YOLOv9s, YOLOv10n, YOLOv10s, and RT-DETR.


\subsection{Evaluation Metrics}
Model performance was evaluated using precision, recall, mAP@0.5, and mAP@0.5:0.95. Precision measures the proportion of correct detections, while recall measures the ability to identify relevant objects. The mAP@0.5 metric evaluates detection accuracy at an IoU threshold of 0.5, whereas mAP@0.5:0.95 averages performance across multiple IoU thresholds. Computational efficiency was evaluated using inference latency, model parameter size, and floating-point operations (FLOPs). These metrics assess the suitability of each model for real-time, vehicle-mounted animal detection.

\section{Results and Discussion} \label{Results and Discussion}

\subsection{Overall Detection Performance}

\begin{table}[!ht]
\caption{Detection performance comparison of object detection models across synthetic and real thermal datasets.}
\centering
\begin{tabular}{p{0.13\columnwidth} p{0.18\columnwidth} p{0.11\columnwidth} p{0.14\columnwidth} p{0.11\columnwidth} p{0.09\columnwidth}}
\Xhline{1.1pt}

\textbf{Model} &
\textbf{Dataset} &
\textbf{$mAP_{0.5}$} &
\textbf{$mAP_{0.5:0.95}$} &
\textbf{Precision} &
\textbf{Recall} \\

\hline

YOLOv8n & Synthetic & 0.9318 & 0.6788 & 0.7857 & 0.9662 \\
YOLOv8n & Real Thermal & 0.9827 & 0.9493 & 0.8384 & 0.9140 \\

YOLOv8s & Synthetic & 0.8837 & 0.6821 & 0.8922 & 0.8263 \\
YOLOv8s & Real Thermal & 0.9874 & 0.9527 & 0.9374 & 0.9120 \\

YOLOv9t & Synthetic & 0.8683 & 0.6596 & 0.9070 & 0.8658 \\
YOLOv9t & Real Thermal & 0.9831 & 0.9483 & 0.8890 & 0.8876 \\

YOLOv9s & Synthetic & 0.8946 & 0.6932 & 0.8623 & 0.8392 \\
YOLOv9s & Real Thermal & 0.9749 & 0.9529 & 0.9074 & 0.9020 \\

YOLOv10n & Synthetic & \textbf{0.7972} & \textbf{0.6092} & 0.8219 & 0.7681 \\
YOLOv10n & Real Thermal & 0.9603 & 0.9399 & 0.9063 & 0.8782 \\

YOLOv10s & Synthetic & 0.9433 & 0.6988 & 0.8098 & 0.9061 \\
YOLOv10s & Real Thermal & \textbf{0.9879} & \textbf{0.9571} & 0.9310 & 0.9189 \\

RT-DETR & Synthetic & 0.9613 & 0.7404 & 0.9219 & 0.9639 \\
RT-DETR & Real Thermal & 0.9841 & 0.9401 & 0.9273 & 0.9608 \\

\Xhline{1.1pt}
\end{tabular}

\label{tab:detection_performance}

\end{table}

The results in Table \ref{tab:detection_performance} demonstrate that models trained on augmented real thermal images consistently achieve the highest overall accuracy. However, models trained exclusively on synthetic thermal data also achieve strong detection performance, confirming the effectiveness of the proposed RGB-to-thermal translation framework.
Notably, RT-DETR achieved the highest synthetic-domain performance with an mAP@0.5 of 0.9613, indicating that the generated thermal images preserve substantial detection-relevant information despite being synthetically produced. A larger performance gap is observed between mAP@0.5 and mAP@0.5:0.95 for models trained on synthetic thermal data compared to those trained on real thermal data. This suggests that while synthetic images effectively preserve object presence, shape, and localization cues required for successful detection at lower IoU thresholds, they do not always retain the fine-grained boundary information necessary for precise object localization at stricter IoU thresholds. In contrast, real thermal images contain authentic thermal characteristics and object contours, resulting in more accurate bounding box regression and consequently higher mAP@0.5:0.95 scores.
\subsection{Effectiveness of Synthetic Thermal Data}
This study evaluates whether synthetic thermal data can serve as an alternative to real infrared imagery for animal detection. Although models trained on real thermal images achieved the highest overall performance, synthetic thermal data produced competitive results across multiple architectures. YOLOv10s achieved 0.9433 mAP@0.5 using only synthetic thermal data, while RT-DETR achieved the best synthetic-domain performance with 0.9613 mAP@0.5. YOLOv8n also achieved high recall of 0.9662, indicating strong detection sensitivity. RT-DETR’s strong performance may be attributed to its transformer-based design, which captures global context and long-range dependencies through self-attention \cite{carion2020detr}. Its multi-scale feature interaction and hybrid encoding strategies further improve object representation and localization \cite{zhao2024rtdetr}. These results suggest that the synthetic thermal images preserve key animal contours, spatial structures, and localization cues, reducing dependence on costly real thermal datasets.

\subsection{Comparison Across Detection Architectures}
Clear performance differences were observed among architectures.
YOLOv8 and YOLOv10 generally provided the best balance between detection accuracy, inference speed, and computational efficiency. YOLOv10s achieved the highest overall performance on the augmented real thermal dataset with mAP@0.5 = 0.9879 and mAP@0.5:0.95 = 0.9571. RT-DETR achieved strong detection performance, particularly on synthetic thermal data, but incurred significantly higher latency and parameter counts compared to YOLO models. These results indicate that transformer-based architectures can achieve strong performance on synthetic thermal data but may be less suitable for latency-constrained deployment due to their higher computational cost. The nano and tiny variants demonstrated lower computational cost and latency, making them attractive for embedded systems despite slightly reduced accuracy.

\subsection{Computational Efficiency}

\begin{table}[!ht]
\caption{Computational efficiency comparison of the detection models.}
\centering
\begin{tabular}{p{0.22\columnwidth} p{0.22\columnwidth} p{0.22\columnwidth} p{0.17\columnwidth}}
\Xhline{1.1pt}

\textbf{Model} &
\textbf{Parameters (M)} &
\textbf{Latency (ms)} &
\textbf{GFLOPs} \\

\hline
YOLOv8n  & 3.0  & 8.37--14.07  & 8.1 \\
YOLOv8s  & 11.1 & 8.58--14.25  & 28.4 \\
YOLOv9t  & 2.0  & 14.44--20.25 & 7.6 \\
YOLOv9s  & 7.2  & 15.23--20.87 & 26.7 \\
YOLOv10n & 2.3  & 8.91--14.78  & 6.5 \\
YOLOv10s & 7.2  & 9.11--14.95  & 21.4 \\
RT-DETR  & 32.0 & 17.92--24.75 & 103.4 \\
\Xhline{1.1pt}
\end{tabular}

\label{tab:computational_efficiency}
\end{table}

Table~\ref{tab:computational_efficiency} summarizes the computational efficiency of the evaluated models. YOLOv8n achieved the lowest inference latency, followed by YOLOv10n, making these models suitable for real-time deployment. Although RT-DETR achieved strong accuracy, its substantially larger parameter count and higher latency may limit deployment feasibility in resource-constrained wildlife monitoring edge systems.

\subsection{Discussion and Future Work}

The experimental results demonstrate that synthetic thermal imagery can provide useful training data for animal detection under low-visibility conditions. Although models trained on augmented real thermal data achieved the highest overall detection accuracy, models trained exclusively on synthetic thermal images remained competitive across multiple architectures. In particular, RT-DETR achieved the highest performance on synthetic thermal data with an mAP@0.5 of 0.9613 and an mAP@0.5:0.95 of 0.7404, while YOLOv8n achieved a recall of 0.9662. These results indicate that the RGB-to-thermal translation process preserves important object-level characteristics, including animal shape, spatial structure, and localization cues required for detection. However, the larger gap between mAP@0.5 and mAP@0.5:0.95 observed for models trained on synthetic data suggests that the generated thermal images may preserve coarse object structure more effectively than the fine-grained boundary information required for precise localization at stricter IoU thresholds. In contrast, models trained on real thermal imagery achieved substantially higher mAP@0.5:0.95 values, suggesting that authentic thermal characteristics and object contours remain beneficial for accurate bounding-box localization. 

The results also reveal an important trade-off between detection performance and computational efficiency. While RT-DETR demonstrated strong detection performance, particularly on synthetic thermal data, its larger parameter count and higher inference latency make it less attractive for resource-constrained deployment. In comparison, lightweight YOLO variants, particularly YOLOv8n and YOLOv10n, achieved substantially lower inference latency while maintaining competitive detection performance, indicating their potential for real-time embedded applications. Nevertheless, the reported latency measurements were obtained on NVIDIA A100 GPUs and therefore should not be interpreted as direct evidence of real-time performance on automotive edge hardware. In addition, the relatively small size of both datasets and differences in class composition between the synthetic and real thermal datasets limit direct cross-domain comparison and may affect the generalizability of the reported results. Despite these limitations, the findings demonstrate that synthetic thermal generation can reduce dependence on scarce annotated infrared imagery and provide a scalable source of training data for low-visibility wildlife detection.

Future work will extend the framework using larger and more diverse synthetic and real thermal datasets and evaluate its performance on independent thermal datasets. This evaluation will assess generalization across diverse animal species, environmental conditions, sensor characteristics, object scales, and imaging scenarios. Additional experiments will investigate whether increasing the diversity and quantity of synthetic thermal images can reduce the localization gap observed at stricter IoU thresholds. Hybrid training strategies that jointly leverage synthetic and real thermal data will also be explored, including different data-mixing, pretraining, and fine-tuning configurations, to improve robustness and reduce the domain gap between generated and real thermal imagery. Finally, future work will focus on deployment-oriented evaluation on embedded automotive platforms such as NVIDIA Jetson Orin and Jetson Xavier. These experiments will characterize inference latency, throughput, memory consumption, and computational resource utilization under resource-constrained conditions. Model optimization techniques may also be investigated to improve deployment efficiency while preserving detection accuracy. Ultimately, evaluation in real-world vehicle-mounted scenarios under nighttime and low-visibility conditions will be necessary to assess the practical reliability and feasibility of the framework for wildlife detection and collision-mitigation applications.
\section{Conclusion}
This study investigates synthetic thermal image generation as a scalable strategy for animal detection under low-visibility conditions. Using CycleGAN-Turbo, RGB wildlife images were translated into synthetic thermal representations and evaluated with multiple YOLO variants and RT-DETR. The results show that synthetic thermal data can preserve detection-relevant features and achieve competitive performance, while models trained on augmented real thermal data achieved the highest overall accuracy. Among the evaluated architectures, lightweight YOLO models provided a favorable balance between detection performance and inference efficiency, whereas RT-DETR achieves strong accuracy at a higher computational cost. Overall, the findings demonstrate the potential of synthetic thermal imagery to reduce reliance on scarce infrared datasets and support the development of efficient vehicle-mounted wildlife detection systems.

\section*{Acknowledgments}
The authors acknowledge the use of generative AI tools exclusively to enhance grammatical accuracy and manuscript readability. These AI tools did not generate original content or findings beyond editorial refinements.

\section*{Conflicts of interest}
The authors declare no conflict of interest.
\bibliographystyle{IEEEtranDOIandURLwithDate}
\bibliography{ref}

\end{document}